\documentclass[conference]{IEEEtran}
\IEEEoverridecommandlockouts
\usepackage{cite}
\usepackage{times}
\usepackage{latexsym}
\usepackage{amsmath,amssymb,amsfonts}
\usepackage{algorithmic}
\usepackage{graphicx}
\usepackage{textcomp}
\usepackage{xcolor}
\usepackage{amsmath}
\usepackage{amsthm}
\usepackage{mathtools}
\usepackage{amsfonts}
\usepackage{framed}
\usepackage{multirow}
\usepackage{float}
\usepackage{subcaption}
\usepackage{microtype}
\def\BibTeX{{\rm B\kern-.05em{\sc i\kern-.025em b}\kern-.08em
    T\kern-.1667em\lower.7ex\hbox{E}\kern-.125emX}}
\DeclareMathOperator*{\rms}{RMS}

\begin{document}

\title{A Topological Approach to Compare Document Semantics Based on a New Variant of Syntactic N-grams}

\author{\IEEEauthorblockN{Fanchao Meng}
\IEEEauthorblockA{\textit{Biocomplexity Institute} \\
\textit{University of Virginia}\\
mf3jh@virginia.edu}}

\maketitle

\begin{abstract}
This paper delivers a new perspective of thinking and utilizing syntactic n-grams (sn-grams). Sn-grams are a type of non-linear n-grams which have been playing a critical role in many NLP tasks. Introducing sn-grams to comparing document semantics thus is an appealing application, and few studies have reported progress at this. However, when proceeding on this application, we found three major issues of sn-grams: lack of significance, being sensitive to word orders and failing on capture indirect syntactic relations. To address these issues, we propose a new variant of sn-grams named generalized phrases (GPs). Then based on GPs we propose a topological approach, named DSCoH, to compute document semantic similarities. DSCoH has been extensively tested on the document semantics comparison and the document clustering tasks. The experimental results show that DSCoH can outperform state-of-the-art embedding-based methods. 
\end{abstract}

%

\section{Introduction}
In this paper, we primarily propose a new variant of sn-grams (viz., Generalized Phrase; GP) and a document semantics comparison method (Document Similarity based on Cohomology; DSCoH) based on GPs and algebraic topology. Most of state-of-the-art sn-grams are defined on directed paths or subgraphs of dependency parse trees \cite{sidorov2019syntactic}. Such definitions are weak at reflecting the significance of a sn-gram in conveying semantics, detecting semantically similar sn-grams with different directions of relations, and capturing indirect syntactic relations. These issues motivate the invention of GPs. To verify the effectiveness of GP, we apply it in a fundamental NLP task, the document semantics comparison problem. Based on GPs we designed DSCoH, a document semantic similarity method, which utilizes algebraic topology techniques \cite{hatcher2002algebraic}. 
\par
A GP is essentially an undirected and weighted sub-tree. Based on this concept, comparing semantics of two sentences can be studied over a graph built upon a \textit{constituency parse tree} (CPT) \cite{jurafsky2014speech} pair and lexical similarity relations interconnecting the trees. Since graphs are considered as \textit{complexes}\footnote{Intuitively, complexes can be understood as objects which ``look like" geometric objects as well as having algebraic characteristics.} \cite{ghrist2014elementary} in algebraic topology, then techniques such as \textit{cohomology} \cite{munkres1984elements} can be utilized to address the problem. This directly motivates our design of DSCoH. We formulate the computation of document semantic similarities as a multi-objective optimization problem. With the help from cohomology, we found that basic cycles imply semantically similar 2-word GPs (i.e. those containing 1 or 2 words), and thus can be used to form approximate solutions. DSCoH implements this idea and computes document similarities. Additionally, we also discuss how the cases with $K$-word GPs can be solved by generalizations of the 2-word GP case. Since DSCoH is based on GPs which are actual constituents, DSCoH is completely explainable, which is superior to many other existing methods. 
\par
To verify their effectiveness of DSCoH, we conducted experiments on the document semantics comparison problem and the document clustering problem, and compared the performance of DSCoH and a set of embedding-based methods. The results are mostly positive for DSCoH especially on actual documents (i.e. those containing multiple sentences).
\par
This paper is organized as follows: Section \ref{sec:gp} reviews some previous work on sn-grams, and discusses how the GP is defined; Section \ref{sec:dscoh} explains the analysis of document semantic similarities from the cohomology perspective, and elaborates DSCoH; and Section \ref{sec:experiments} shows all experiments and results.

\section{A Variant of Syntactic N-grams}\label{sec:gp}

\subsection{Previous Work \& Issues}
Sn-grams are defined to be a type of non-linear n-grams, and are typically constructed from dependency parse trees, each consisting of a path or a sub-tree \cite{sidorov2014syntactic, sidorov2019syntactic}. Sn-grams have been shown to be useful in various NLP tasks such as authorship attribution \cite{sidorov2013syntactic, sidorov2014syntactic, posadas2015syntactic}, machine translation \cite{sennrich2015modelling}, dependency parsing \cite{ng2015web}, contextual polarity analysis \cite{agarwal2009contextual} and language modeling \cite{wu1999combining}.
\par

When trying to utilize current sn-grams in comparing document semantics, we found three major issues: 
\par
\begin{figure}
	\centering
	\includegraphics[width=0.5\textwidth]{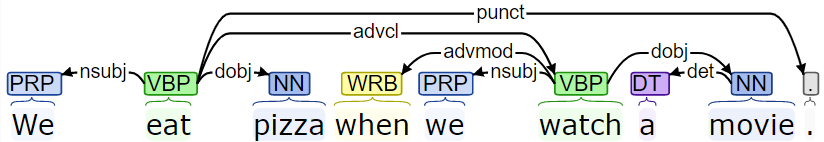}
	\caption{``\textit{Eat}" and ``\textit{movie}" have an indirect dependency relation, and the path length reflects that the relation is not immediate but is merely 1 greater than ``\textit{watch}" and ``\textit{movie}".}
	\label{fig:sngram_sig}
\end{figure}
\begin{figure}
	\centering
	\includegraphics[width=0.5\textwidth]{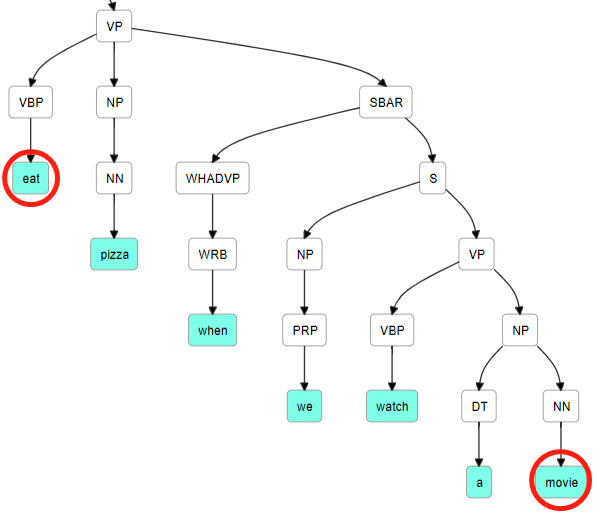}
	\caption{The path length between ``\textit{Eat}" and ``\textit{movie}" in the CPT more effectively reflects the weakness of their relation.}
	\label{fig:sngram_sig_2}
\end{figure}

First, a syntactic n-gram needs to be considered its significance in representing semantics. Path lengths in parse trees provide an assessment of the strength of the relationship between words in a sentence. However, dependency parse trees have some limitations compared to constituency-based parse trees. For example, consider the sentence: ``\textit{We eat pizza when we watch a movie.}", in which ``\textit{eat}" and ``\textit{movie}" do have an (indirect) syntactic relation yet much weaker than that between ``\textit{watch}" and ``\textit{movie}". Figures \ref{fig:sngram_sig} and \ref{fig:sngram_sig_2} visualize these relations, and the path lengths in CPT can reflect the significances of n-grams more effectively.
\par
\begin{figure}
	\centering
	\includegraphics[width=0.35\textwidth]{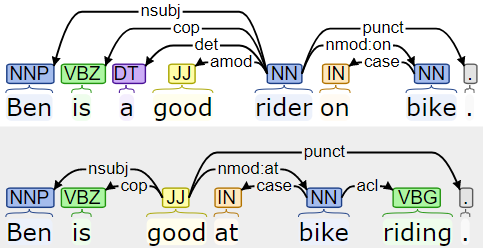}
	\caption{``\textit{Rider}" and ``\textit{bike}" are actually semantically similar to ``\textit{bike}" and ``\textit{riding}".From the perspective of semantic relationships, dependency-based parse trees can be too sensitive to word order.}
	\label{fig:word_order_2}
\end{figure}

Second, sn-grams are too sensitive to word order. For example, consider the two sentences in Figure \ref{fig:word_order_2}, in which ``\textit{rider}" and ``\textit{bike}" do not have significant difference in semantics from ``\textit{bike}" and ``\textit{riding}" but they are assigned with opposite directions in the dependency parse trees. On the other hand, constituency parse trees do not rely on directional syntactic relations. 
\par
Third, dependency parse tree based sn-grams can only capture immediate dependencies between words rather than indirect syntactic relations, while constituency parse trees are natural and straightforward on both direct and indirect syntactic relations. 
\par

\subsection{Generalized Phrases}
Based on the preceding discussion, we propose a variant of the sn-grams, named the \textbf{generalized phrase (GP)}, and defined below. GPs captures the significance of sn-grams in reflecting semantics, are not sensitive to word order, and are able to capture indirect syntactic relations. 
\begin{framed}
	\noindent\underline{\textbf{Definition: Generalized Phrase (GP)}}
	\par
	A \textbf{generalized phrase} is a minimal non-empty subtree of the CPT\footnotemark containing at least one leaf. The leaves are considered orderless. 
\end{framed}
\footnotetext{\textit{Stopwords}~\cite{onixstopwords2} should always be removed and should never appear in any \textit{GP}. The parse trees will be pruned if necessary. 
}
Note that the significance of the relatedness between the leaves is determined by the tree structure excluding the leaves. The simpler the structure, the more significant the GP. For 2-word GPs the significance is computed by using the path length between the two leaves, and for $K$-word GPs there can be multiple approaches to compute the significance, for example, considering the average of path lengths of all pairs of leaves. 

\section{Document Semantics Comparison}\label{sec:dscoh}

\subsection{Motivations and Methods}
We begin by explaining the motivations for DSCoH. First, we formulate the problem of computing the semantic similarity between two sentences\footnote{The similarity between two documents can be straightforwardly computed by the sum of all sentence similarities.} as a multi-objective optimization problem; and second, we propose a framework based on cohomology theory \cite{hatcher2002algebraic} producing approximate solutions to the optimization problem efficiently.
\par
\noindent\textbf{Semantic Similarity as an Optimization Problem:}
\par
Considering the GP concept introduced above, to compute the semantic similarity between two sentences, a general idea is to find as many as possible semantically similar GPs across the two sentences. Thus, maximizing the accumulation of similarities contributed by such semantically similar GPs is a core requirement to compute the sentence similarity, which one of our objectives. However, on the other hand, not all GPs are significant as discussed above. Thus, we also need to keep the GPs taken the accumulation as significant as possible, which is the other objective. Formally, let $s_i$ and $s_j$ denote two input sentences; let $\mathbb{GP}_i$ and $\mathbb{GP}_j$ denote the set of all GPs in consideration in the two sentences; let $Sim: \mathbb{GP}_i \times \mathbb{GP}_j \rightarrow \mathbb{R}_{\geq 0}$ denote a bounded and real-valued function computing the similarity between two GPs, the greater the output value, the more similar the GPs; let $Sig: \mathbb{GP} \rightarrow \mathbb{R}_{> 0}$ denote a bounded and real-valued function computing the significance weight of a GP, the greater the output value, the more significant the input GP; let $Sig_m$ denote the supremum of the set of possible significance values; and let $RMS$ denote the \textit{root mean square} function. Note that $Sim$ may also need to take the significance weights of GPs into consideration. Then the optimization problem is formulated as follows:
\begin{align*}
	& \max\limits_{\mathbb{A} \subseteq \mathbb{GP}_i \times \mathbb{GP}_j} \sum\limits_{(GP_x, GP_y) \in \mathbb{A}}Sim(GP_x, GP_y) \\
	& \min\limits_{\mathbb{A} \subseteq \mathbb{GP}_i \times \mathbb{GP}_j} \sum\limits_{(GP_x, GP_y) \in \mathbb{A}}  \rms\Big((Sig(GP_x) - Sig_{m}), \\
	& (Sig(GP_y) - Sig_m)\Big) \\
	& \text{s.t. } \mathbb{GP}_i \times \mathbb{GP}_j < \infty
\end{align*}
The semantic similarity between $s_i$ and $s_j$ finally is computed by $\sum\limits_{(GP_x, GP_y) \in \mathbb{A}} Sim(GP_x, GP_y)$. 
\par
\noindent\textbf{Framework with Cohomology:}
\par
This problem is NP-hard\footnote{The hardness of this problem can be proved by using tree homomorphism problem \cite{hell1996complexity} and general multi-objective optimization problems \cite{glasser2010approximability}. We skip the proof as it is not a primary concern in this paper.}, and the search space can be exponential (which is justified below). Next, we show how this problem can be naturally understood and solved from the cohomology theory\footnote{Readers who are interested in algebraic topology are referred to \cite{munkres1984elements, hatcher2002algebraic, edelsbrunner2010computational}. Through this paper, most algebraic topology concepts are used without formal definitions but references are always provided.} perspective, and a framework solving the problem is proposed.
\par
\begin{figure}
	\centering
	\includegraphics[width=0.25\textwidth]{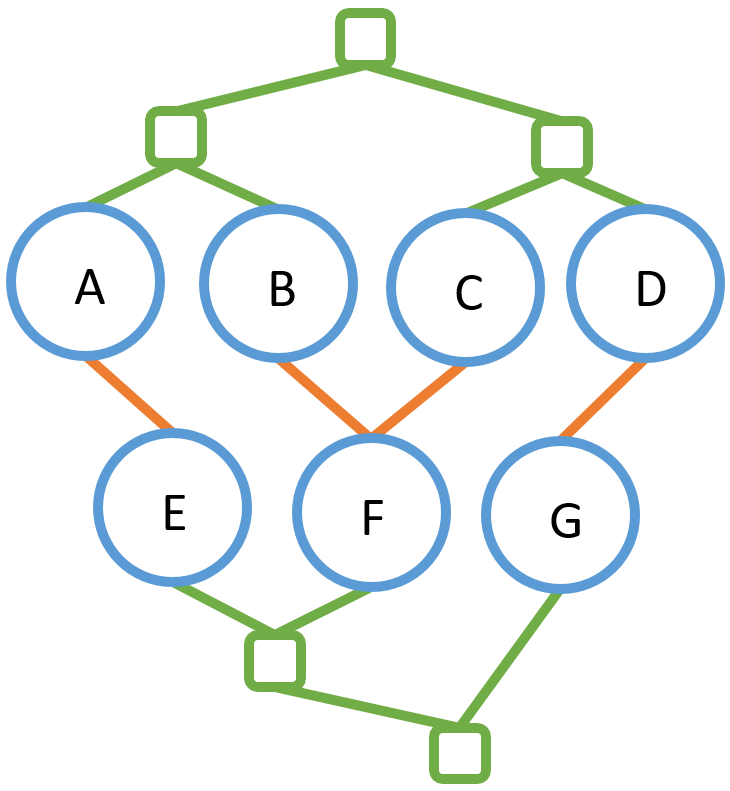}
	\caption{An example of syn-sim graph. A, B, C, D, E, F and G represent words. Green links represent syntactic relations, and orange links represent lexical similarities in consideration.}
	\label{fig:syn_sim_graph}
\end{figure}

We start with a special yet typical case of the problem: seeking a set $\mathbb{A}$ containing only 2-word GPs\footnote{Empirically, for convenience, in some cases we expand a single word into a 2-word GP consisting of two same words with the path length $1$ if necessary.} which produces an optimal solution. As to $Sim$ and $Sig$, the similarities between GPs primarily come from lexical similarities, and the significance weights of GPs come from syntactic relations between words. Thus, it is intuitive to have these information in a single object which by our design is an undirected and weighted graph. The construction of this graph is as follows. First, given two sentences, a CPT is computed for each of them\footnote{In the implementation, stop words are removed and the trees are pruned (See Section \ref{sec:algorithms}).}. The weight on each edge in the parse trees is set to $1$. We call the edges in parse trees \textbf{syn-edges}. Second, we compute the lexical similarity (a real value) between each word in a sentence and every word in the other. By applying a predetermined threshold to the similarities, an edge is created between each word pair, across the two sentences, which has a similarity greater than or equal to the threshold, and the weight on this edge is assigned the value of similarity. We call these edges \textbf{sim-edges}. Finally, we union the two parse trees and the set of edges created by lexical similarities, and produce the desired graph. We call this graph the \textbf{syn-sim graph} \footnote{The name, \textit{syn-sim}, means ``syntactic relations" and ``similarities".} of the two sentences. An example of syn-sim graph is shown in Figure \ref{fig:syn_sim_graph}.
\par
Given the syn-sim graph, the candidates for a solution are straightforward. Since trees are acyclic, then every pair of sim-edges induces a cycle in the graph which implies a pairing of two semantically similar GPs. Such GP pairs are candidate elements in a solution. However, in the worst case, there are $mn$ sim-edges, where $m$ and $n$ are the numbers of leaves in the two parse trees; $mn \choose 2$ GP pairs; and thus $O(2^{mn \choose 2})$ possible solutions in the search space. Therefore, finding ``good" cycles efficiently for the objective functions is the core task in solving the optimization problem, which is the place where the cohomology theory can play a role as cycles in a graph are low dimensional ``holes" in a topological space and the cohomology theory concentrates on finding and describing such ``holes" \cite{edelsbrunner2010computational, munkres1984elements, hatcher2002algebraic}. 
\par

Cohomology is one of the primary perspectives in algebraic topology, and it is convenient in handling pure algebraic objects. By construction, syn-sim graphs are troublesome as a geometric object because of the coexistence of two different types of edges (i.e., syn-edges and sim-edges). Thus, in our analysis syn-sim graphs are considered as algebraic objects (as discussed below), and we utilize cohomology to study cycles in the syn-sim graphs. 
\par
To work with cohomology, an \textit{abstract simplicial complex}\footnote{An \textit{abstract simplicial complex} is a collection $\mathcal{S}$ of finite sets, such that if $\mathcal{A}$ is an element of $\mathcal{S}$, so is every subset of $\mathcal{A}$.} \cite{munkres1984elements} (which is an algebraic object) is required. The construction is as follows. Given a syn-sim graph $\mathcal{G}$, we contract each sim-edge to a super-vertex. For example, in Figure \ref{fig:syn_sim_graph}, the super-vertices can be $X_1=(A,E)$, $X_2=(B,F)$, $X_3=(C,F)$ and $X_4=(D,G)$. Then, the set of all such super-vertices forms an abstract simplicial complex, $\mathcal{S}$. $\mathcal{G}$ with sim-edges substituted by super-vertices can be considered as a \textit{geometric realization} \cite{munkres1984elements} of $\mathcal{S}$. For convenience, we define the group of \textit{$p$-cochains} \cite{munkres1984elements} for the abstract simplicial complex to be $Hom(\mathcal{S}, \mathbb{R}_{\geq 0})$ which is a \textit{functor} \cite{munkres1984elements} containing all homomorphisms of $\mathcal{S}$ into $\mathbb{R}_{\geq 0}$. Particularly, $Sim \in Hom(\mathcal{S}, \mathbb{R}_{\geq 0})$ for $0$-cochains. Additionally, $\mathbb{Z}_2$ is used as the ground field for coefficients of $p$-cochains. Based on these settings, we propose three lemmas which show that it is feasible to solve the optimization problem with approximation by utilizing cohomology. 
\par
\noindent\textbf{Lemma \#1:} \textbf{$\sum\limits_{i} c^0([X_i])$ is a \textit{$0$-cocycle} but is not a \textit{$0$-coboundary} (i.e. $\sum\limits_{i} c^0([X_i])$ is generator of the \textit{cohomology group} $H^0(\mathcal{S})$) \textmd{\cite{munkres1984elements}}, where $c^0 \in Hom(\mathcal{S}, \mathbb{R}_{\geq 0})$ denotes elementary $0$-cochain, and $[X_i]$ denotes $0$-simplex.} This holds as $\mathcal{G}$ is connected \cite{hatcher2002algebraic}. 
\par
\begin{figure}
	\centerline{\includegraphics[width=0.5\textwidth]{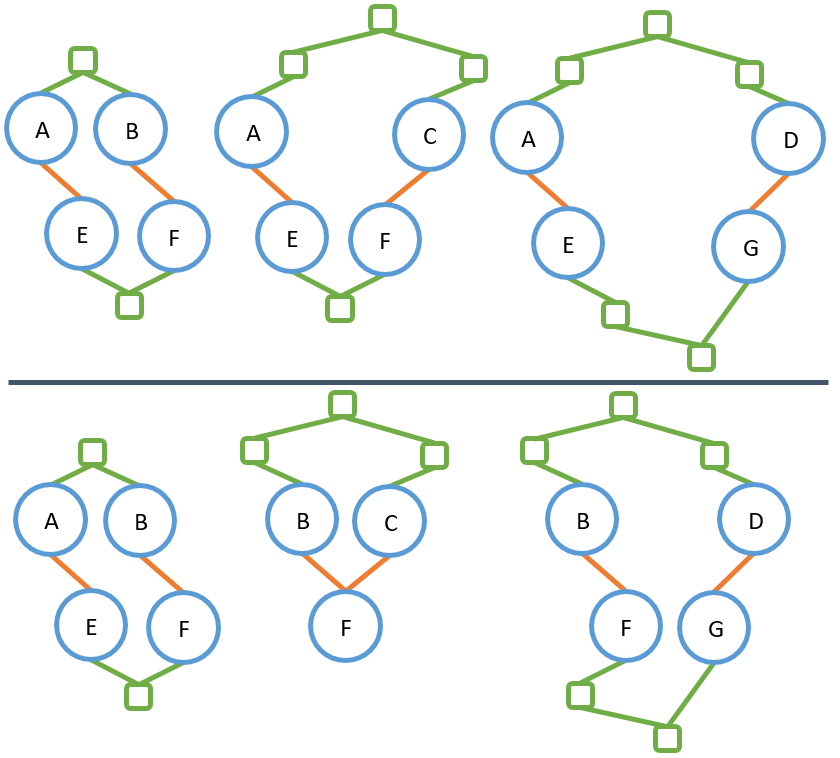}}
	\caption{Two examples of cycle bases for the graph in Figure \ref{fig:syn_sim_graph}. The upper basis is obtained from $\delta(c^0[X_1]) = c^1([X_1, X_2]) + c^1([X_1, X_3] + c^1([X_1, X_4]))$, and the sum of weights is 17 (Recall that the sim-edges have been contracted to a super-vertex when computing coboundaries.). The lower basis is obtained from $\delta(c^0[X_1]) = c^1([X_1, X_2]) + c^1([X_1, X_3] + c^1([X_1, X_4]))$, and the sum of weights is 15, which makes it the minimum cycle basis.}	
	\label{fig:cycle_basis}
\end{figure}

\noindent\textbf{Lemma \#2:} \textbf{For each elementary $0$-cochain, its \textit{$1$-coboundary} is a cycle basis of $\mathcal{G}$.} To prove this lemma, it is sufficient to show that, first,the elementary $1$-cochains are cycles in $\mathcal{G}$; second, the elementary $1$-cochains are independent of each other; and third, the $1$-coboundary contains $e - v + c$ (the circuit rank \cite{berge2001theory}) elementary $1$-cochains, where $e$ denotes the number of edges in $\mathcal{G}$, $v$ denotes the number of vertices, and $c$ denotes the number of connected components. Given the construction of $\mathcal{S}$, the first condition automatically holds. The second condition holds because elementary cochains are defined to be independent \cite{munkres1984elements}. To prove the third condition, we let $e_{sim}$ denote the number of sim-edges; and let $e_i$ denote the number of incident syn-edges of the vertices. Additionally, since $\mathcal{G}$ is always connected, then $c = 1$. Also, since for a tree the number of vertices is always one more than the number of edges, and $\mathcal{G}$ contains two parse trees, then $e_i = v - 2$. Thus, $e - v + c = (e_i + e_c) - v + 1 = e_c - 1$. Again, since $\mathcal{G}$ is always connected, then all $X_i$'s are reachable from each other. Thus, by the definition of the \textit{coboundary operator} \cite{munkres1984elements}, $\delta (c^0([X_i])) = \sum\limits_{i \sim j} c^1([X_i, X_j])$, where $\delta$ is the coboundary operator and $i \sim j$ denotes adjacency, it always holds that $\delta (c^0([X_i]))$ has $e_c - 1$ elementary $1$-cochains. Figure \ref{fig:cycle_basis} shows two examples of this lemma.
\par
\noindent\textbf{Lemma \#3:} \textbf{The set of elementary $1$-cochain contained in all $\delta (c^0([X_i]))$ contains all basic cycles.} The proof of this lemma can be done by way of contradiction. If a basic cycle was not contained, then it could be contained in $\delta (c^0([X_i]))$ for any $X_i$, which implies that this basic cycle had to be disconnected from all $X_i$. This contradicts the connectivity of $\mathcal{G}$.
\par
\begin{figure}
	\centerline{\includegraphics[width=0.5\textwidth]{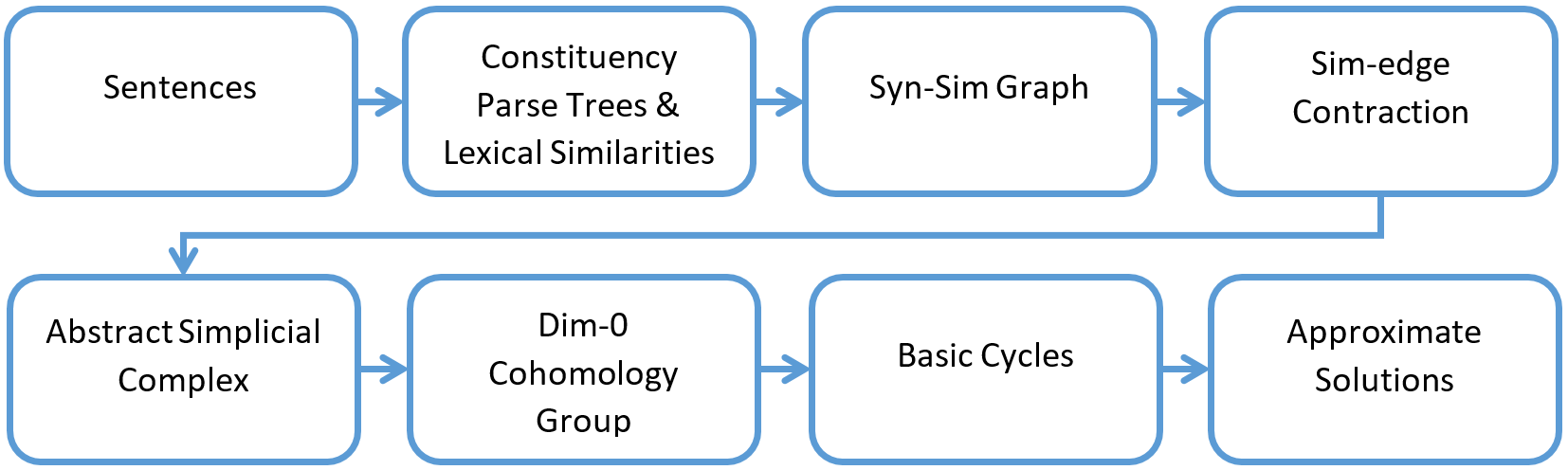}}
	\caption{The framework with cohomology.}
	\label{fig:cohomology_framework}
\end{figure}

The three lemmas above imply that each $\delta (c^0([X_i]))$ is an approximate solution to the optimization problem, and no candidate pair of GPs to the solutions is missed. Thus, finding the best solutions from basic cycles of $\mathcal{G}$ can be expected to have fairly good approximate solutions, for example, solving the minimum cycle basis problem on $\mathcal{G}$. Figure \ref{fig:cohomology_framework} summarizes the framework with cohomology. The time complexity of this framework is dominated by the one that is of the higher order between computing constituency parse trees and computing desired cycle basis. Theoretically, for the former task, methods based on the \textit{Cocke–Younger–Kasami algorithm} can be higher than $O(s^3 |G|)$, where $s$ is the sentence length and $G$ is the CNF grammar \cite{hopcroft2008introduction}. For the latter task, the state-of-the-art method proposed in \cite{mehlhorn2009minimum} runs in $O(b^2a/ loga+ba^2)$, where $a$ is the number of vertices and $b$ is the number of edges, and thus in our case it is $O(a^3)$.
\par
\noindent\textbf{Generalization to $K$-word GPs:}
\par
It has been shown that a pairing of 2-word GPs is induced by two sim-edges in a syn-sim graph. Analogously, a pairing of $K$-word GPs is induced by $K$ sim-edges\footnote{Similar to 2-word GPs, when considering $K$-word GPs, in some cases, we need to expand a single word into a chain of copies of the word with each edge in the chain being assigned a weight $1$.}. Thus, it is straightforward to induce from \textbf{Lemma \#2} that a pairing of $K$-word GPs always corresponds to $(K-1)$ basic cycles, and naturally such a pairing can be expressed as a formal sum of the $(K-1)$ basic cycles. This conclusion directly suggests an approach to compute semantic similarity between two sentences by utilizing $K$-word GPs. That is, first, we solve the optimization problem with $2$-word GPs and obtain a cycle basis (e.g. a minimum cycle basis); second, we collect all pairings of $K$-word GPs, and for each pairing we decompose it into a subset of basic cycles; and finally, we sum the similarity results contributed by all $K$-word GP pairings and produce a solution. The time complexity of this method is the running time of the framework in the 2-word GP case plus $O({x \choose K})$, where $x$ is the number of sim-edges. However, if a minimum cycle basis is utilized, a subset of basic cycles may not form a minimum cycle basis for the subgraph induced by a $K$-word GP pairing. Thus, a better solution is that for each $K$-word GP pairing we induce the subgraph from the pairing, then solve the optimization problem on this subgraph in the 2-word GP case, and finally sum the solutions to all sub-problems. The time complexity of this refined method is dominated by the higher order one between computing CPTs and $O({x \choose K}K^3))$. It can be concluded from these generalization methods that a generalization may run much slower than the 2-word GP case and it may not gain much benefit in comparing semantics of sentences as the fundamental elements that contribute to the sentence similarities are still pairings of 2-word GPs (i.e. basic cycles). Therefore, we suggest to use the 2-word GP case to compute sentence similarities, and we leave the study of more advanced generalization methods to future work.
\par
Next, we propose a concrete algorithm to compute the semantic similarity between two documents utilizing this framework in the 2-word GP case.

\subsection{Algorithms} \label{sec:algorithms}
In this section, we propose an algorithm computing the semantic similarity between two documents. We name it \textbf{DSCoH}\footnote{\textit{DSCoH} is short for ``document similarity based on cohomology".}. DSCoH is an implementation of the framework with cohomology considering only 2-word GPs. This algorithm solves the minimum cycle basis problem to obtain pairings of semantically similar GPs. To simplify syn-sim graphs, we also propose a tree pruning algorithm and integrate it into DSCoH. DSCoH also contains a designed $Sim$ function taking lexical similarities, GP significances and sentence lengths into consideration. We elaborate on DSCoH below.
\par

\begin{framed}
	\noindent\underline{\textbf{Algorithm 1: DSCoH}}
	\par
	\noindent\textbf{Given:}\par
	$\bullet$ Two documents, $D_i$ and $D_j$.
	\par
	$\bullet$ A threshold for lexical similarities, $\theta_w$.
	\par
	\noindent\textbf{Seek:}\par
	A real value as the semantic similarity between $D_i$ and $D_j$, denoted by $\varphi_D(D_i, D_j)$.
	\par
	\noindent\textbf{Stage 1:} \textbf{Constituency Parse Trees} 
	\par
	For each document $D_i$ and for each sentence $S_{ik} \in D_i$, where $k$ indexes the sentences, compute a \textit{CBPT} for $S_{ik}$, denoted by $T_{ik}$.
	\par
	\noindent\textbf{Stage 2:} \textbf{Pruned Parse Trees}
	\par
	For each $T_{ik}$, apply \textbf{Algorithm 2} to prune $T_{ik}$. The pruned tree is denoted by $\hat{T_{ik}}$.
	\par
	\noindent\textbf{Stage 3:} \textbf{Lexical Similarity Relations}
	\par
	For each pruned tree pair $(\hat{T_{ik}}, \hat{T_{jh}})$, where $\hat{T_{ik}}$ and $\hat{T_{jh}}$ are in $D_i$ and $D_j$ respectively, identify all word pairs $\{(t_{ik}^a, t_{jh}^b)\}$, where $t_{ik}^a \in \hat{T_{ik}}$ and $t_{jh}^b \in \hat{T_{jh}}$, with lexical similarities $\varphi_w(t_{ik}^a, t_{jh}^b) \geq \theta_w$.
	\par
	\noindent\textbf{Stage 4:} \textbf{Syn-Sim Graphs}
	\par
	For each pair of pruned trees $\hat{T_{ik}}$ and $\hat{T_{jh}}$, union the trees and create an edge for each identified word pair obtained from \textbf{Stage 3} to construct a syn-sim graph, denoted by $\mathcal{G}_{ik,jh}$.
	
	\par
	\noindent\textbf{Stage 5:} \textbf{Minimum Cycle Basis}
	\par
	For each syn-sim graph $\mathcal{G}_{ik,jh}$ over $k$ and $h$, set the weight on each sim-edge to $\infty$ (only for this stage) and compute a minimum cycle basis, $\mathbb{B}_{kh}$.
	\par
	\noindent\textbf{Stage 6:} \textbf{GP Significance Weights}
	\par
	For each basic cycle $c_l = \{GP_{l1}, GP_{l2}\}$, compute a weight for $c_l$ based on the path lengths of $GP_{l1}$ and $GP_{l2}$, called the \textbf{GP significance weight}, denoted by $w_{c_l}$:
	\begin{equation*}
		w_{c_l} = \frac{\theta_{c1}}{p_{l1}^{\theta_{c2}} + p_{l2}^{\theta_{c2}}}
	\end{equation*}
	where $\theta_{c1}$ and $\theta_{c2}$ are predetermined parameters which empirically are set to 3, $p_{l1}$ and $p_{l2}$ are path lengths of $GP_{l1}$ and $GP_{l2}$. 
	\par
	\noindent\textbf{Stage 7:} \textbf{Sentence Length Weights}
	\par
	For each pruned tree pair $(\hat{T_{ik}}, \hat{T_{jh}})$, compute a weight, $w_{s_{kh}}$, reflecting how much the basic cycles have covered the semantics of the two sentences. The weight is called the \textbf{sentence length weight}, and computed by
	\begin{equation*}
		w_{s_{kh}} = \frac{2 |\mathbb{B}_{kh}|}{{L_{ik} \choose 2} + |L_{ik}| + {L_{jh} \choose 2} + |L_{jh}|}
	\end{equation*}
	where $L_{ik}$ and $L_{jh}$ denote the numbers of leaves in $\hat{T_{ik}}$ and $\hat{T_{jh}}$ respectively.
	\par
	\noindent\textbf{Stage 8:} \textbf{Document Semantic Similarity}
	\par
	The similarity between two GPs in a basic cycle is computed by
	\begin{equation*}
		\varphi_c(c_l) = w_{c_l} \cdot \min\Big(\varphi_w(t_k^a, t_h^a), \varphi_w(t_k^b, t_h^b)\Big)
	\end{equation*}
	The similarity between $D_i$ and $D_j$ thus is computed by
	\begin{equation*}
		\varphi_D(D_i, D_j) = \sum\limits_{k,h} \Bigg(w_{s_{kh}} \cdot \sum\limits_{c_l \in \mathbb{B}_{kh}} \varphi_c(c_l)\Bigg)
	\end{equation*}
\end{framed}
\begin{framed}
	\noindent\underline{\textbf{Algorithm 2: Parse Tree Pruning}}
	\par
	\noindent\textbf{Given:}
	\par
	$\bullet$ A CPT $T$.
	\par
	$\bullet$ A \textit{stopword} list, $\mathbb{W}_s$.
	\par
	$\bullet$ A set of \textit{name entities} in consideration, $\mathbb{N}$.
	\par
	$\bullet$ A set of \textit{POS} tags in consideration, $\mathbb{P}$.
	\par
	\noindent\textbf{Seek:}
	\par
	A pruned tree.
	\par
	\noindent\textbf{IF} $T$ is a single-node tree, and the only node is denoted by $t_0$:
	\par
	\noindent\qquad\textbf{IF} $t_0 \in \mathbb{W}_s$ \textbf{OR} $t_0 \notin \mathbb{N}$ \textbf{OR} $t_0$ is a punctuation:
	\par
	\noindent\qquad\qquad\textbf{RETURN} An empty tree.
	\par
	\noindent\qquad\textbf{ELSE}:
	\par
	\noindent\qquad\qquad\textbf{RETURN} $T$.
	\par
	\noindent\textbf{ELSE}:
	\par
	\noindent\qquad\textbf{IF} the root $r$ of $T$, such that $r \notin \mathbb{P}$:
	\par
	\noindent\qquad\qquad\textbf{RETURN} An empty tree.
	\par
	\noindent\qquad\textbf{ELSE IF} $r$ has only one child $c_0$:
	\par
	\noindent\qquad\qquad Apply \textbf{Algorithm 2} to the subtree rooted at $c_0$.
	\par
	\noindent\qquad\qquad\textbf{RETURN} The pruned subtree rooted at $c_0$.
	\par
	\noindent\qquad\textbf{ELSE}:
	\par
	\noindent\qquad\qquad Prune all subtrees rooted at $r$ by applying \textbf{Algorithm 2} recursively. 
	\par
	\noindent\qquad\qquad Remove from $r$ the subtrees which have empty pruned trees. 
	\par
	\noindent\qquad\qquad\textbf{RETURN} The pruned tree, $\hat{T}$.
	
\end{framed}
As to \textbf{Algorithm 1}, a number of comments are added here: 
\par
First, to compute lexical similarities there are multiple approaches, for example, approaches based on word sense disambiguation such as ADW \cite{pilehvar2013align} and approaches using word embedding models such as LexVec \cite{salle2016matrix}. 
\par
Second, in \textbf{Stage 5}, the weights on sim-edges are set to $\infty$ because doing this will force a basic cycle can only contain 2 sim-edges, which keeps every basic cycle corresponding to a pair of semantically similar GPs. Also, to compute a minimum cycle basis\footnote{Note that in general minimum basis cycles in a graph may not be unique.}, we utilize the method described in \cite{kavitha2008tilde}.
\par
Third, a significance weight measures the significance of the similarity between two paired GPs in reflecting the document similarity. This weight should be induced by the function $Sig$, and in our case $Sig(GP_x)$ is determined by the path length of $GP_x$. A significance weight is high only when the paired GPs are both significant. 
\par
Fourth, since DSCoH concentrates on ``what are similar" rather than ``what are not similar", then it happens that, for example, suppose that sentences $S_1$ and $S_2$ have the same GP pairings to $S_1$ and $S_3$, even though $S_3$ may be much longer than $S_2$, then in this case DSCoH will anyway give the same similarity between $S_1$ and $S_2$ as that between $S_1$ and $S_3$, though in fact $S_1$ and $S_3$ are very likely to be less similar than the other pair. The sentence length weights are introduced to offset this bias. 
\par
Fifth, DSCoH follows a modular design. Several components such as $w_{c_l}$, $w_{s_{kh}}$ and $\varphi_c$ can be substituted by other implementations. 
\par
As to \textbf{Algorithm 2}, $\mathbb{W}_s$, $\mathbb{N}$ and $\mathbb{P}$ all act as filters to rule out constituents not in consideration, and can be customized. The final pruned tree is in the simplest form. 

\section{Experiments}\label{sec:experiments}
To verify the effectiveness of DSCoH, we test it into two tasks, the document semantics comparison and the document clustering. In the document semantics comparison task, DSCoH performs as well as, sometimes better than, other methods. In the document clustering task, DSCoH outperforms almost all other methods. We elaborate on each of the tasks below. 

\subsection{Document Semantics Comparison}\label{sec:document_semantics_comparison}
This task requires each method to take every pair of documents in a dataset and produce a value as the similarity. These similarity values are compared to a set of values determined by human judges. The Spearman correlation for each method is then computed to measure how close to the human judges this method performs. 
\par
\noindent\textbf{Datasets:}
\par
The datasets in use include: \textbf{Lee60} \cite{lee2005empirical} which contains 60 document pairs\footnote{In \cite{lee2005empirical}, the original dataset consists of 50 documents (i.e. 1225 document pairs). However, the distribution of human judge scores in the original dataset is extremely skewed. To balance the bias, we applied a systematic sampling method \cite{thompson2012sampling} and obtained 60 pairs for our experiments.}, \textbf{Li30} \cite{li2006sentence} which contains 30 sentence pairs, \textbf{STS2017} \cite{cer2017semeval} which contains 250 sentence pairs, and \textbf{SICK} \cite{marelli-etal-2014-sick} which contains 9840 sentence pairs. 
\par
\noindent\textbf{Methods to Compare:}
\par
A set of pretrained embedding model based methods are used for comparison including \textbf{Word2Vec} \cite{mikolov2013distributed} equipped with \textbf{WMD} \cite{kusner2015doc}, \textbf{NASARI}\footnote{The word embedding model trained on the UMBC corpus is used.} \cite{camacho2015nasari}, \textbf{Doc2Vec} \cite{le2014distributed}, \textbf{GloVe} \cite{pennington2014glove}, \textbf{fastText} \cite{bojanowski2017enriching}, \textbf{LexVec} \cite{salle2016matrix} and \textbf{Sent2Vec} \cite{pagliardini2017unsupervised}. Each document is represented as a vector by utilizing the embedding models, and then we use Cosine similarity to compute a similarity value for every pair of documents. Note that Doc2Vec and Sent2Vec can directly take a document as input and produce a vector. For others, we use the conventional average word vector method to represent documents.  
\par
Our method is DSCoH utilizing NASARI for lexical similarities. Note that DSCoH can interface with any lexical similarity method, and NASARI is randomly selected without any particular purpose. Additionally, since the lexical similarity threshold is a parameter for DSCoH, we test 10 settings and show the results.
\par
\noindent\textbf{Experimental Results:}
\par
The results are shown in Table \ref{tab:spearman}.
\begin{table}
	\centering
	\resizebox{0.8\columnwidth}{!}{
		\begin{tabular}{|c|c|c|c|c|}
			\hline
			\textbf{}            & \multicolumn{4}{c|}{\textbf{Spearman Correlation}}                \\ \hline
			\textbf{Methods}     & \textbf{Lee60} & \textbf{Li30} & \textbf{STS2017} & \textbf{SICK} \\ \hline
			\textbf{DSCoh-0.1} & 0.56           & 0.58          & 0.10             & 0.22          \\ \hline
			\textbf{DSCoh-0.2} & 0.64           & 0.70          & 0.31             & 0.37          \\ \hline
			\textbf{DSCoh-0.3} & 0.73           & 0.73          & 0.45             & 0.44          \\ \hline
			\textbf{DSCoh-0.4} & 0.79           & 0.82          & 0.56             & 0.47          \\ \hline
			\textbf{DSCoh-0.5} & 0.82           & \textbf{0.86} & 0.63             & 0.48          \\ \hline
			\textbf{DSCoh-0.6} & \textbf{0.85}           & 0.80          & 0.66             & 0.49          \\ \hline
			\textbf{DSCoh-0.7} & 0.82           & 0.71          & 0.68             & 0.49          \\ \hline
			\textbf{DSCoh-0.8} & 0.77           & 0.68          & 0.70             & 0.50          \\ \hline
			\textbf{DSCoh-0.9} & 0.77           & 0.66          & 0.70             & 0.50          \\ \hline
			\textbf{DSCoh-1.0} & 0.77           & 0.66          & 0.70             & 0.50          \\ \hline
			\textbf{Doc2Vec}     & 0.57           & 0.78          & 0.75             & 0.56          \\ \hline
			\textbf{NASARI}  & 0.79           & 0.83          & 0.71             & 0.55          \\ \hline
			\textbf{GloVe}   & 0.81           & 0.67          & 0.72             & 0.54          \\ \hline
			\textbf{WMD}         & 0.82  & 0.78          & \textbf{0.80}    & 0.57          \\ \hline
			\textbf{LexVec}      & 0.77           & 0.72          & 0.73             & \textbf{0.61} \\ \hline
			\textbf{fastText}    & 0.71           & 0.79          & 0.72             & 0.54          \\ \hline
			\textbf{Sent2Vec}    & 0.83           & 0.82          & 0.74             & 0.55          \\ \hline
		\end{tabular}
	}
	\caption{Spearman correlations for Lee60, Li30, STS2017 and SICK. The best score for each category is bold.}
	\label{tab:spearman}
\end{table}
\par
\noindent\textbf{Discussion:}
\par
Two important observations are discussed here. One is that DSCoH is relatively stable as the similarity threshold varies, which makes DSCoH friendly in real practice. The other is that DSCoH performs better on actual documents (i.e. those containing several sentences) than sentences. It obtains a good performance on Li30 because Li30 was created by using words' interpretation sentences in a dictionary. Those sentences are mostly short and contain few expressive words. Consequently the basic cycles, in a comparison, captured by DSCoH would have carried almost the whole semantics if the two sentences are similar. In STS2017 and SICK, sentences are typically longer than those in Li30. Thus, DSCoH may not capture as adequate semantics in two sentences as embedding based methods. On the other hand, for actual documents, DSCoH would be able to capture adequate key semantics while embedding based methods typically would have taken ``too much" semantics into consideration such that some unimportant semantics become noise. We justify this claim in Section \ref{sec:document_clustering}.

\subsection{Document Clustering}\label{sec:document_clustering}
This task requires each method to group a set of documents by their semantics. The grouping results then are compared to the human judgments by computing \textit{Adjusted Rand Index} (ARI) \cite{hubert1985comparing}, \textit{Normalized Mutual Information} (NMI) \cite{vetterling1992numerical} and \textit{Fowlkes-Mallows Index} (FMI) \cite{fowlkes1983method}. All of these scores range in $[0, 1]$. They measures how well a clustering produced by a method matches the human judgments, the higher the better.
\par
\noindent\textbf{Datasets:}\footnote{Datasets are attached as supplementary materials.}
\par
Three datasets are in use including \textbf{20Newsgroups} \cite{lang1995newsweeder}, \textbf{Reuters-21578} \cite{lewis1987reuters} and \textbf{BBC} \cite{greene2006practical}. Based on 20Newsgroups, two sampled datasets are created: \textbf{20News-M5} containing 5 categories without confusion and \textbf{20News-C10} containing 10 categories with confusion. Based on Reuter-21578, a sampled set is created: \textbf{Reuters-M7} containing 7 categories. Based on BBC, a sampled set is created: \textbf{BBC-M5}. In each category in the four sampled sets, 50 documents are sampled. 
\par
\noindent\textbf{Methods to Compare:}
\par
The methods for comparison are the same as those in Section \ref{sec:document_semantics_comparison} except WMD, and the clustering method is spectral clustering \cite{ng2002spectral} \footnote{Note that there may be more state-of-the-art methods specific to document clustering; however, the main objective of this section is to verify the effectiveness of DSCoH and GP rather than focusing on clustering problems.}.
\par
\noindent\textbf{Experimental Results:}
\par
The results are shown in Table \ref{tab:document_clustering}. 
\begin{table}
	\begin{subtable}{0.5\textwidth}
		\resizebox{1\columnwidth}{!}{
			\begin{tabular}{|c|c|c|c|c|c|c|c|}
				\hline
				\multicolumn{8}{|c|}{\textbf{20News-M5}}                                                                                     \\ \hline
				\textbf{}    & \textbf{DSCoH} & \textbf{Doc2vec} & \textbf{NASARI} & \textbf{fastText} & \textbf{Sent2Vec} & \textbf{LexVec} & \textbf{GloVe} \\ \hline
				\textbf{ARI} & \textbf{0.90}  & 0.86             & 0.79            & 0.64              & 0.56              & 0.85            & 0.80           \\ \hline
				\textbf{NMI} & \textbf{0.91}  & 0.84             & 0.81            & 0.68              & 0.67              & 0.85            & 0.83           \\ \hline
				\textbf{FMI} & \textbf{0.92}  & 0.89             & 0.83            & 0.71              & 0.67              & 0.88            & 0.84           \\ \hline
			\end{tabular}
		}
		\label{tab:20news_m5_k5}
	\end{subtable}
	
	\begin{subtable}{0.5\textwidth}
		\resizebox{1\columnwidth}{!}{
			\begin{tabular}{|c|c|c|c|c|c|c|c|}
				\hline
				\multicolumn{8}{|c|}{\textbf{20News-C10}}                                                                                    \\ \hline
				\textbf{}    & \textbf{DSCoH} & \textbf{Doc2Vec} & \textbf{NASARI} & \textbf{fastText} & \textbf{Sent2Vec} & \textbf{LexVec} & \textbf{GloVe} \\ \hline
				\textbf{ARI} & \textbf{0.64}  & 0.47             & 0.57            & 0.48              & 0.34              & 0.63            & 0.53           \\ \hline
				\textbf{NMI} & \textbf{0.73}  & 0.56             & 0.66            & 0.60              & 0.48              & 0.72            & 0.65           \\ \hline
				\textbf{FMI} & \textbf{0.68}  & 0.52             & 0.61            & 0.53              & 0.42              & 0.67            & 0.58           \\ \hline
			\end{tabular}
		}
		\label{tab:20news_c10_k10}
	\end{subtable}
	
	\begin{subtable}{0.5\textwidth}
		\resizebox{1\columnwidth}{!}{
			\begin{tabular}{|c|c|c|c|c|c|c|c|}
				\hline
				\multicolumn{8}{|c|}{\textbf{Reuters-M7}}                                                                                    \\ \hline
				\textbf{}    & \textbf{DSCoH} & \textbf{Doc2vec} & \textbf{NASARI} & \textbf{fastText} & \textbf{Sent2Vec} & \textbf{LexVec} & \textbf{GloVe} \\ \hline
				\textbf{ARI} & \textbf{0.90}  & 0.40             & 0.61            & 0.52              & 0.75              & 0.58            & 0.64           \\ \hline
				\textbf{NMI} & \textbf{0.91}  & 0.52             & 0.67            & 0.62              & 0.79              & 0.64            & 0.69           \\ \hline
				\textbf{FMI} & \textbf{0.92}  & 0.49             & 0.66            & 0.59              & 0.78              & 0.64            & 0.69           \\ \hline
			\end{tabular}
		}
		\label{tab:reuters_m7_k7}
	\end{subtable}
	
	\begin{subtable}{0.5\textwidth}
		\resizebox{1\columnwidth}{!}{
			\begin{tabular}{|c|c|c|c|c|c|c|c|}
				\hline
				\multicolumn{8}{|c|}{\textbf{BBC-M5}}                                                                                        \\ \hline
				\textbf{}    & \textbf{DSCoH} & \textbf{Doc2Vec} & \textbf{NASARI} & \textbf{fastText} & \textbf{Sent2Vec} & \textbf{LexVec} & \textbf{GloVe} \\ \hline
				\textbf{ARI} & 0.88           & 0.55             & 0.72            & 0.60              & \textbf{0.89}     & 0.82            & 0.80           \\ \hline
				\textbf{NMI} & 0.86           & 0.57             & 0.73            & 0.64              & \textbf{0.88}     & 0.81            & 0.80           \\ \hline
				\textbf{FMI} & 0.90           & 0.64             & 0.78            & 0.68              & \textbf{0.91}     & 0.85            & 0.84           \\ \hline
			\end{tabular}
		}
		\label{tab:bbc_m5_k5}
	\end{subtable}
	\caption{Scores for document clustering tasks.}
	\label{tab:document_clustering}
\end{table}
\par
\noindent\textbf{Discussion:}
\par
The results show that DSCoH outperforms almost all methods except Sent2Vec, though they are very close. These results strongly support the claim proposed in Section \ref{sec:document_semantics_comparison}. Thus, it is optimistic to conclude that DSCoH would perform well in document-based tasks.

\section{Conclusion \& Future Work}
Several takeaways are: first, GPs are effective in representing document semantics; second, DSCoH is effective in comparing document semantics; third, DSCoH is completely explainable and fourth, algebraic topology techniques are not adornments but insightful tools in comparing document semantics.
\par
On the other hand, the work in this paper still has a huge room to grow. First, could we design a single document semantic representation? Second, are $K$-GPs unnecessary or we have not understood them well? Third, computing constituency parse trees is always a pain as to running time. Could we use dependency parse trees instead? And fourth, could we make DSCoH more effective to sentences?


\bibliographystyle{mybst}
\bibliography{mybib}

\end{document}